\documentclass[11pt]{article}

\usepackage[final]{acl}

\usepackage{times}
\usepackage{latexsym}

\usepackage[T1]{fontenc}

\usepackage[utf8]{inputenc}

\usepackage{microtype}

\usepackage{inconsolata}

\usepackage{graphicx}
\usepackage{fancyvrb}
\usepackage{framed}
\usepackage{lineno}
\usepackage{hyperref}
\usepackage{url}
\usepackage{booktabs}       
\usepackage{amsfonts}       
\usepackage{nicefrac}       
\usepackage{microtype}      
\usepackage{xcolor}         
\usepackage{xspace}
\usepackage{amsmath}
\usepackage{amssymb}
\usepackage{multirow}
\usepackage{multicol}
\usepackage{booktabs}
\usepackage[export]{adjustbox}
\usepackage{fontawesome5}
\usepackage{hyperref}
\usepackage[noabbrev,capitalize,nameinlink]{cleveref}
\usepackage{listings}
\usepackage[linesnumbered, ruled, vlined]{algorithm2e}
\usepackage[most]{tcolorbox}
\usepackage{wrapfig}
\usepackage{subcaption}
\usepackage[
ragged
]{sidecap}

\tikzstyle{rq-box} = [draw=white, fill=blue!5, rounded corners, 
    rectangle, inner sep=7pt, inner ysep=10pt, 
    text width=0.4\textwidth, align=justify]
\tikzstyle{rqbox-title} =[fill=blue!20, text=black, rounded corners]

\tcbset{
    promptstyle/.style={
        enhanced,
        width=\linewidth,
        size=small,
        colback=white,
        colframe=black,
        fonttitle=\footnotesize,
        colbacktitle=gray!20,
        coltitle=black,
        rounded corners,
        boxrule=0.5pt,
        drop shadow=black!50!white,
        attach boxed title to top left={
            xshift=-2mm,
            yshift=-2mm
        },
        boxed title style={
            rounded corners,
            size=small, 
            colback=gray!20
        }
    }
}

\newtcolorbox{promptbox}[1][]{
    promptstyle,
    title=Prompt,
    #1
}

\newcommand{\rone}{\texttt{Deepseek-R1}}
\newcommand{\qwq}{\texttt{QwQ-32B}}
\newcommand{\dataset}{\textbf{fs1}}
\newcommand{\datasetrt}{\textbf{\textbf{rt}}}

\newcommand*\circled[1]{\tikz[baseline=(char.base)]{
            \node[shape=circle,draw,inner sep=.6pt] (char) {#1};}}

\newcommand{\up}[1]{{\color{blue} $\uparrow${#1}}}

\title{Follow the Path: Reasoning over Knowledge Graph Paths\\ to Improve Large Language Model Factuality}

\author{Mike Zhang\textsuperscript{{$\diamond$}{$\ddagger$}{$\dagger$}} \hspace{2em} Johannes Bjerva\textsuperscript{$\ddagger$} \hspace{2em} Russa Biswas\textsuperscript{$\ddagger$}\\
  \textsuperscript{$\diamond$}University of Copenhagen\\
  \textsuperscript{$\ddagger$}Aalborg University \\
  \textsuperscript{$\dagger$}Pioneer Centre for Artificial Intelligence\\
  \texttt{mike.zhang@di.ku.dk}\hspace{2em} \texttt{\{jbjerva, rubi\}@cs.aau.dk}\\}

\begin{document}
\maketitle
\begin{abstract}

We introduce \dataset{}, a simple yet effective method that improves the factuality of reasoning traces by collecting them from large reasoning models and grounding them in knowledge graph (KG) paths.
We fine-tune eight instruction-tuned Large Language Models (LLMs) on 3.9K factually grounded reasoning traces and rigorously evaluate them on six complex open-domain question-answering (QA) benchmarks encompassing 23.9K questions.
Our results demonstrate that our \dataset-tuned model consistently outperforms instruction-tuned counterparts with parallel sampling by 6-14 absolute points (pass@$16$).
Our detailed analysis shows that \dataset{} considerably improves model performance over more complex questions (requiring 3 or more hops on KG paths) and numerical answer types compared to the baselines. 
Furthermore, in single-pass inference, we notice that smaller LLMs show the most improvements.
While prior works demonstrate the effectiveness of reasoning traces primarily in the STEM domains, our work shows strong evidence that anchoring reasoning to factual KG paths is a critical step in transforming LLMs for reliable knowledge-intensive tasks.
\end{abstract}

\begin{figure}[t]
    \centering
    \includegraphics[width=\linewidth]{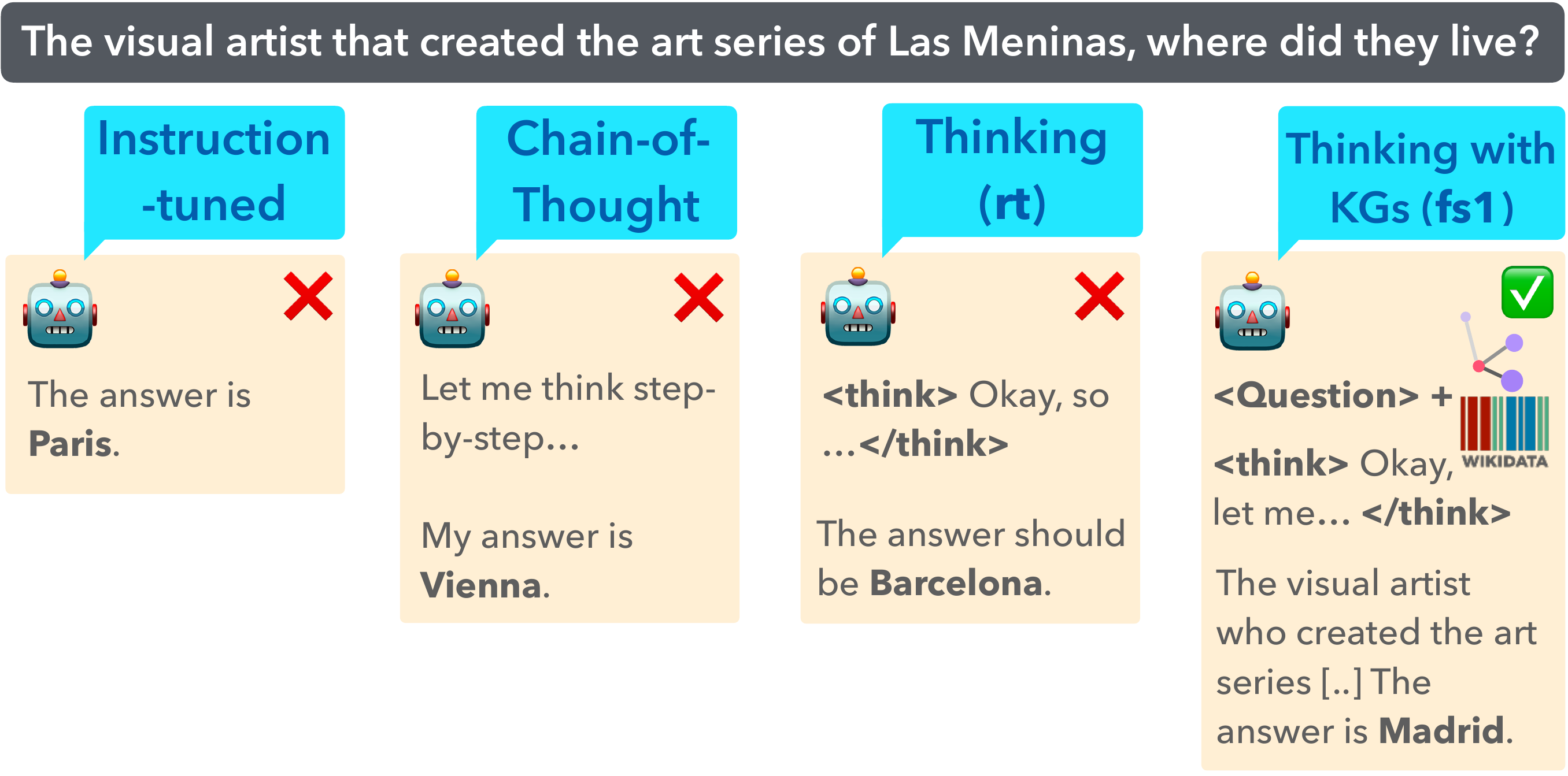}
    \caption{\textbf{Snapshot of Method.} We show a snapshot of the experiments executed in this study. There are four settings on how a question can be answered; (1) direct answer from an instruction-tuned model, (2) step-by-step reasoning via Chain-of-Thought, (3) original ``thinking'', and (4) knowledge-graph enhanced ``thinking''. We show an example of (4) in~\cref{fig:fs1}.}
 \label{fig:fig1}
\end{figure}

\section{Introduction}\label{introduction}

Factual consistency of LLM-generated output is a requirement for critical real-world applications.
LLM reasoning in the form of ``thinking'' has shown promising improvements in model performance on complex downstream tasks, such as mathematical reasoning and puzzle-like questions using additional compute resources during inference (e.g., test-time scaling;~\citealp{wu2024inference,muennighoff2025s1simpletesttimescaling,zhang2025whathowwherewell}). 
However, it remains an open question whether these reasoning techniques improve factuality, particularly for complex multi-hop QA (mQA).
This task tests a model's ability to answer a question by synthesizing information from multiple pieces of evidence, often spread across different resources and requiring reasoning steps. 
We hypothesize that reasoning models should perform better than non-reasoning LLMs on the mQA task.
To test this hypothesis, we source reasoning traces from state-of-the-art reasoning models and fine-tune several non-reasoning LLMs to attempt to induce reasoning capabilities.
However, we have no guarantee that these reasoning traces from the large reasoning models are factually correct.
In order to have a formal factual grounding in these traces, we condition the models on retrieved knowledge graph (KG) paths relevant to the questions. 
This is possible as KGs encode facts as directed, labeled graphs over entities and relations, which offers a verifiable foundation to inform each step of the reasoning process.
We call our approach \dataset{} (\emph{factual} simple test-time scaling;~\citealp{muennighoff2025s1simpletesttimescaling}).

We fine-tune eight different LLMs sizes on the original reasoning traces (\textbf{rt}; 3.4K samples) or on our KG-enhanced traces (\dataset{}; 3.9K samples). 
We evaluate the fine-tuned models on six QA test sets spanning 23.9K questions, finding that fine-tuning on this amount of data can improve accuracy by 6-14 absolute points (pass@$16$) for a 32B parameter model across the benchmarks.
A snapshot of our method is in~\cref{fig:fig1}. 
This setup enables us to address our research question (\textbf{RQ}): \emph{To what extent does grounding the reasoning processes of LLMs in KG paths enhance their factual accuracy for mQA?} To address this question, our contributions are:

\begin{itemize}
\itemsep0em
\item[\circled{1}] We demonstrate that with test-time scaling (parallel sampling), our \dataset-tuned Qwen2.5-32B model improves factual accuracy by 6-14 absolute points (at pass@$16$).  
\item[\circled{2}] We conduct an analysis over the question and answer types (e.g., question difficulty, answer type, and domains) to investigate where \dataset{}-tuned models provide improvements. \emph{We find that \dataset-tuned models perform better on more difficult questions, requiring 3 hops or more}.
\item[\circled{3}] We examine performance of eight \dataset-tuned models (360M-32B parameters) in a pass@$1$ setting against baselines.
\emph{We find that smaller LLMs have the largest increase in performance, whereas larger models see less profound improvements}.
\item[\circled{4}] We release 3.4K raw reasoning traces and 3.9K KG-enhanced reasoning traces both sourced from \qwq{} and \rone{}.\footnote{Code, datasets, and models are publicly available: \url{https://github.com/jjzha/fs1} (\texttt{MIT}).}
\end{itemize}

\section{Reasoning Data}\label{reasoning-data}

\paragraph{\datasetrt{}: Distilling Reasoning Traces.}
To obtain reasoning traces, we use \texttt{ComplexWebQuestions} (CWQ;~\citealp{talmor-berant-2018-web}), a dataset designed for complex mQA. 
The CWQ dataset is created by automatically generating complex SPARQL queries based on Freebase~\citep{10.1145/1376616.1376746}. 
These queries are then automatically transformed into natural language questions, which are further refined by human paraphrasing. 
We take the CWQ dev.\ set, which consists of 3,519 questions, to curate the reasoning traces.
We query both \qwq{}~\citep{qwq32b} and \rone{} (685B;~\citealp{deepseekai2025deepseekr1incentivizingreasoningcapability}).
By querying the model directly with a question, e.g., ``{\it What art movement was Pablo Picasso part of?}'', we retrieve the reasoning traces surrounded by ``think'' tokens (\texttt{<think>...</think>}) and force the model to give the final answer to the question in \verb$\boxed{}$ format. We extract 3.4K correct-only traces (final answer is correct), which we call \datasetrt{}. We show full examples in~\cref{fig:ex-reasoning-trace-r1} and \cref{fig:ex-reasoning-trace-qwq} (\cref{app:example}).

\paragraph{\dataset{}: Enhancing Reasoning Traces with Knowledge Graph Paths.}
We attempt to steer the reasoning traces with KG paths to remove the inaccuracies in the traces. 
Since the CWQ dataset consists of entities from Freebase, we align them to their corresponding Wikidata entities. Our method extracts all minimum-hop paths between the question entities and the gold answer entities in the KG. Specifically, we first search for 1-hop paths, and if any exist, we stop; otherwise, we search for 2-hop paths, then 3-hop paths, and so on. This minimum-hop constraint serves as an implicit semantic filtering mechanism, as shorter paths in knowledge graphs are generally more likely to represent direct and meaningful relationships. When multiple entities are present in the question, we perform path extraction in two ways: (1) querying each question entity individually with the answer entity, and (2) querying all question entities jointly with the answer entity in a single SPARQL query to capture paths involving multiple question entities. Similarly, when a question has multiple gold answers, each question–answer combination is queried accordingly. The explicit SPARQL queries are in~\cref{app:sparql}. 
For each question in the development set of the CWQ dataset, relevant KG paths are extracted from Wikidata using random walks using SPARQL queries as previously mentioned. 
Each mQA pair in the dataset may contain multiple valid KG paths, which are linearized graphs that retain the structural information of the KG. The paths are generated by extracting the relevant entities from the question and the gold answer. These diverse KG paths that can lead to the same answer reflect the possible diversity of the reasoning traces. Therefore, including linearized graphs improves the interpretability and the explainability of the reasoning traces.
We use the ``subject, relation, object'' format because it is the standard representation of knowledge graph triples and preserves the correct semantic direction of relations.
The prompt to obtain the improved reasoning traces is shown in \cref{fig:fs1}.
Full examples are in \cref{fig:ex-reasoning-trace-fs1-qwq} and \cref{fig:ex-reasoning-trace-fs1-rone} (\cref{app:example}).

\paragraph{Data Statistics.} In~\cref{tab:data-statistics} and~\cref{fig:fs1-violin}, we compare the reasoning trace accuracy and statistics of \textbf{rt} and \dataset{}.
We evaluate reasoning traces using three methods: (1) \textit{Exact Match}, checking if the \verb|\boxed{}| answer exactly matches or is a sub-phrase of any gold answer; (2) \textit{Semantic Match}, accepting answers with a cosine similarity score $>$0.5; and (3) \textit{LLM-as-a-Judge}, verifying entity alignment using \texttt{gpt-4o-mini-2024-07-18}. 
Results show that \dataset{} achieves higher accuracy, indicating that it contains more factual answers.
Traces from \textbf{rt} are longer (up to 1K subwords), \dataset{} traces are typically shorter (around 800 subwords). The median length in subwords is similar for \qwq{} (552 for \textbf{rt} and 553 for \dataset{}), while there is a difference for \rone{} (635 median for \textbf{rt} and 496 for \dataset{}). Spot-checking final answers reveals that \dataset{} yields a more definitive answer.

\begin{table}[t]
    \centering
        \resizebox{\linewidth}{!}{%
            \begin{tabular}{l r r r r r r r}
            \toprule
                && \multicolumn{2}{c}{\qwq{}} & \multicolumn{2}{c}{\texttt{R1-685B}} & \multicolumn{2}{c}{\textsc{Total}} \\
                \cmidrule(lr){3-4} \cmidrule(lr){5-6} \cmidrule(lr){7-8}
                && \multicolumn{1}{c}{\textbf{rt}} & \multicolumn{1}{c}{\dataset{}} & \multicolumn{1}{c}{\textbf{rt}} & \multicolumn{1}{c}{\dataset{}} & \multicolumn{1}{c}{\textbf{rt}} & \multicolumn{1}{c}{\dataset{}} \\
            \midrule
            Exact Match                       && 0.46     & \up{0.63} & 0.56     & \up{0.72}  & 0.51     & \up{0.67} \\ 
            Sem. Match (\texttt{MiniLM})      && 0.50     & \up{0.58} & 0.55     & \up{0.63}  & 0.52     & \up{0.60} \\ 
            LLM-as-a-Judge ({\tt 4o-mini})     && 0.44     & \up{0.61} & 0.54     & \up{0.70}  & 0.49     & \up{0.65} \\ 
            \midrule
            \textit{Samples correct answers}  &&          &           &          &            &          & \\ 
            Number of Samples                 && 1,533    & 1,972     & 1,901    & 1,914      & 3,434    & 3,886  \\
            Avg. Reasoning Length             && 937      & 897       & 1,043    & 637        & 990      & 767    \\
            Avg. Answer Length                && 40       & 93        & 64       & 116        & 52       & 104    \\
            \bottomrule
            \end{tabular}%
        }
        \captionof{table}{\textbf{Training Data Statistics.} Statistics of reasoning traces. \dataset{} has higher performance compared to \datasetrt{}.}
        \label{tab:data-statistics}
\end{table}

\begin{figure}[t!]
    \centering
        \begin{tcolorbox}[title=fs1 Prompt Example, promptstyle, width=\linewidth] 
        \lstset{
            basicstyle=\tiny\sffamily\tiny,
            breaklines=true,
            frame=none,
            columns=fullflexible,
        }
        \begin{lstlisting}
When did the sports team owned by Leslie Alexander win the NBA championship?

While answering the question, make use of the following linearised graph as an inspiration in your reasoning, not as the only answer:

1994 NBA Finals, winner, Houston Rockets
Houston Rockets, owned by, Leslie Alexander
1995 NBA Finals, winner, Houston Rockets
Houston Rockets, owned by, Leslie Alexander.

Put your final answer within \boxed{}.
--
(For illustration) Gold Answer: ["1994 and 1995"]
        \end{lstlisting}
        \end{tcolorbox}
        \caption{\textbf{fs1 Prompt Example.} We depict how we prompt both \rone{} and \qwq{} to obtain better reasoning traces with KG paths.}
        \label{fig:fs1}
\end{figure}

\section{Methodology}\label{methodology}

\subsection{Training and Inference}
We fine-tune six \texttt{Qwen2.5-Instruct} models (0.5B to 32B) on \datasetrt{} and \dataset{}, using only reasoning traces with correct final answers. During inference, we evaluate the model on the original questions to test its performance. Following~\citet{muennighoff2025s1simpletesttimescaling}, we train for 5 epochs with a sequence length of 8,192, a batch size of 16, a learning rate of $1 \times 10^{-5}$ (cosine schedule, 5\% warmup), and a weight decay of $1 \times 10^{-4}$. The models are optimized with a standard supervised fine-tuning (SFT) loss, which minimizes the negative log-likelihood (implemented as the cross-entropy function) of target tokens in an autoregressive manner. Let $y_t^*$ be the correct token and $p_\theta(y_t^* \mid x, y_{<t})$ be the model's probability of predicting it. The model optimizes:
\begin{equation}
\mathcal{L}_\text{SFT}(\theta)=-\frac{1}{T} \sum_{t=1}^T \log p_\theta\left(y_t^* \mid x, y_{<t}\right).
\end{equation}
For inference, we use a temperature ($T$) of $0.7$ and \verb|top_p| of $0.8$ for original instruct models. Otherwise, we use $T = 0.6$ and \verb|top_p| of $0.95$. 
Further details on hardware and costs are in~\cref{app:trainandinference}.

\begin{figure}
        \centering
        \includegraphics[width=\linewidth]{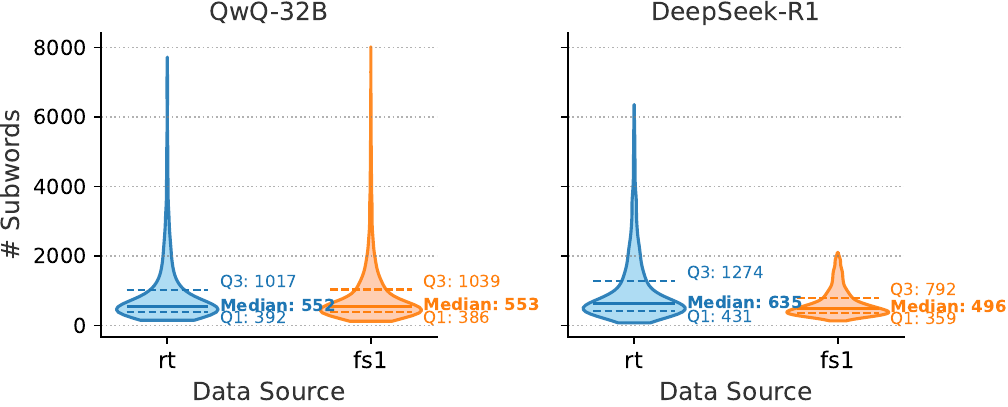}
        \caption{\textbf{Distribution of Reasoning Traces.} Distribution of reasoning length. Particularly for \dataset{} and \rone{}, the reasoning length is shorter.}
        \label{fig:fs1-violin}
\end{figure}

\begin{figure}[t!]
\begin{tcolorbox}[title=LLM-as-a-Judge (Llama-3.3-70B), promptstyle, width=\linewidth]
        \lstset{
            basicstyle=\tiny\sffamily\tiny,
            breaklines=true,
            frame=none,
            columns=fullflexible,
        }
        \begin{lstlisting}[linewidth=\linewidth]
gold answer: ["Joule per gram per kelvin", "Joule per kilogram per kelvin"]
predicted answer: "J/(kg $\cdot$ K)"

Is the gold answer entity or value contained in the predicted answer? Respond only with 0 (no) or 1 (yes).

# Llama-3.3-70B-Instruct outputs "1"
        \end{lstlisting}
        \end{tcolorbox}
        \caption{\textbf{Prompt for LLM-as-a-Judge.} We show the LLM-as-a-Judge prompt for evaluating whether the predicted and gold answer refer to the same real-world entity, where \emph{regular exact string matching will not capture the alignment between the gold and predicted answer} in this example (i.e., the measurement unit).}
        \label{fig:llm-judge}

\end{figure}

\subsection{Benchmarks and Evaluation}\label{subsec:eval}
We show the test datasets, licenses, and in~\cref{tab:qa-datasets}. CWQ is an mQA dataset from WebQuestionsSP with compositional SPARQL queries for Freebase paraphrased by crowd workers. ExaQT is a temporal-QA benchmark combining eight KB-QA datasets, focusing on time-specific queries. GrailQA is a QA dataset with annotated answers and logical forms (SPARQL/S-expressions) across 86 domains. SimpleQA is a fact-seeking QA dataset with verified answers, designed to measure and challenge the factual accuracy of language models. Mintaka is multilingual QA (9 languages), entity-linked pairs across diverse domains, we only take the English split. WebQSP is based on WebQuestions annotated with SPARQL ($\sim$82\% coverage).
We have four baselines, namely \texttt{Qwen2.5-72B-Instruct}~\citep{qwen2.5}, \qwq{}, \rone{}, and \texttt{o3-mini}~\citep{o3-mini}. 
To evaluate our models, we select a suite of six mQA benchmarks with a total of 23.9K questions. 
We have four setups for benchmarking the models: (1) All models including baselines are evaluated zero-shot (i.e., only querying the question); (2) the models are queried using zero-shot chain-of-thought prompting~\citep{kojima2022large, wei2022chain}, where we simply append the prompt ``Put your final answer within \verb|\boxed{}|. Think step-by-step.''; (3) we benchmark the models fine-tuned on \datasetrt{}; (4) we benchmark the models fine-tuned on \dataset{}.
In~\cref{fig:test-examples} (\cref{app:test-examples}), we show an example of each dataset in the test benchmark. 

\begin{table}[t]
    \centering
        \centering
        \scriptsize
            \begin{tabular}{lll}
            \toprule
            \textbf{Dataset} & \textbf{License} & \textbf{Test Size} \\
            \midrule
            CWQ~\citep{talmor-berant-2018-web} 
              & \texttt{apache-2.0}
              & 3.5K\\
            ExaQT~\citep{10.1145/3459637.3482416} 
              & \texttt{cc-by-4.0}
              & 3.2K \\
            GrailQA~\citep{10.1145/3442381.3449992} 
              & \texttt{apache-2.0}
              & 6.8K \\
            SimpleQA~\citep{wei2024measuring} 
              & \texttt{MIT}
              & 4.3K \\
            Mintaka~\citep{sen-etal-2022-mintaka} 
              & \texttt{cc-by-4.0} 
              & 4.0K \\
            WebQSP~\citep{yih-etal-2016-value} 
              & \texttt{apache-2.0}
              & 2.0K \\
              \midrule
              \textsc{Total}& & 23.9K \\
            \bottomrule
            \end{tabular}%
        \captionof{table}{\textbf{Test Benchmark.} Overview of the mQA test sets used in our evaluation.}
        \label{tab:qa-datasets}
\end{table}

\paragraph{Evaluation Metric.}
Similar to previous studies, e.g.,~\citet{ma2025reasoningmodelseffectivethinking}, we report pass@$k$, which reflects the probability that at least one out of $k$ randomly selected completions (drawn from a total of $n$ completions per problem) is correct. As such, it serves as an upper-bound on practical performance, which would require a subsequent selection mechanism. Formally, pass@$k$ is given by: $\mathbb{E}_{\text{problems}}\left[1 - \frac{\binom{n-c}{k}}{\binom{n}{k}}\right]$, where $n$ is the number of generated completions per problem and $c$ is the count of correct completions~\citep{chen2021evaluating}. For our benchmarks, we evaluate $k = \{1, 2, 4, 8, 16\}$. In practice, pass@32 is typically reported for formal theorem-proving tasks, while pass@1 (reducing to standard top-1 accuracy) is standard for math and coding tasks as mentioned by~\citet{ma2025reasoningmodelseffectivethinking}. We report until $k=16$. 

We deliberately evaluate the pass@k upper bound to isolate our method's raw generation and knowledge elicitation capabilities. Introducing a selection mechanism would conflate our generation quality with the accuracy of the specific verifier. Verifiers are inherently bottlenecked by generation recall; they cannot select a correct answer if the model fails to generate it. By significantly expanding the pool of correct reasoning paths, our method solves this critical recall bottleneck.

\paragraph{LLM-as-a-Judge.} To decide whether an answer is correct or not (1 or 0), our main evaluation approach is using LLM-as-a-judge with \texttt{Llama-3.3-70B-Instruct}\footnote{We compare both \texttt{gpt-4o-mini-2024-07-18} and \texttt{Llama-3.3-70B-Instruct} on a subsample of our outputs and saw there is almost no difference in predictions. Additionally, \texttt{Llama-3.3-70B} is rated higher in \href{https://lmarena.ai}{LM Arena} than \texttt{gpt-4o-mini} (at time of writing 79$^\text{th}$ vs. 83$^\text{rd}$ respectively).} to determine whether a predicted answer obtained from the \verb|\boxed{}| output is referring to the same real-world entity as the gold answer. 
An example of this is shown in~\cref{fig:llm-judge}.
When the model does not generate a \verb|\boxed{}| output, we take the last 10 subwords as predicted answer, which LLM-as-a-judge can infer what the predicted real-world entity is when there is not exact string matching.
This same approach is used in~\cref{tab:data-statistics}. Compared to exact string matching and semantic similarity evaluation methods, LLM-as-a-Judge rates the quality of output similarly compared to the other methods.

\begin{figure}[t]
    \centering
    \includegraphics[width=\linewidth]{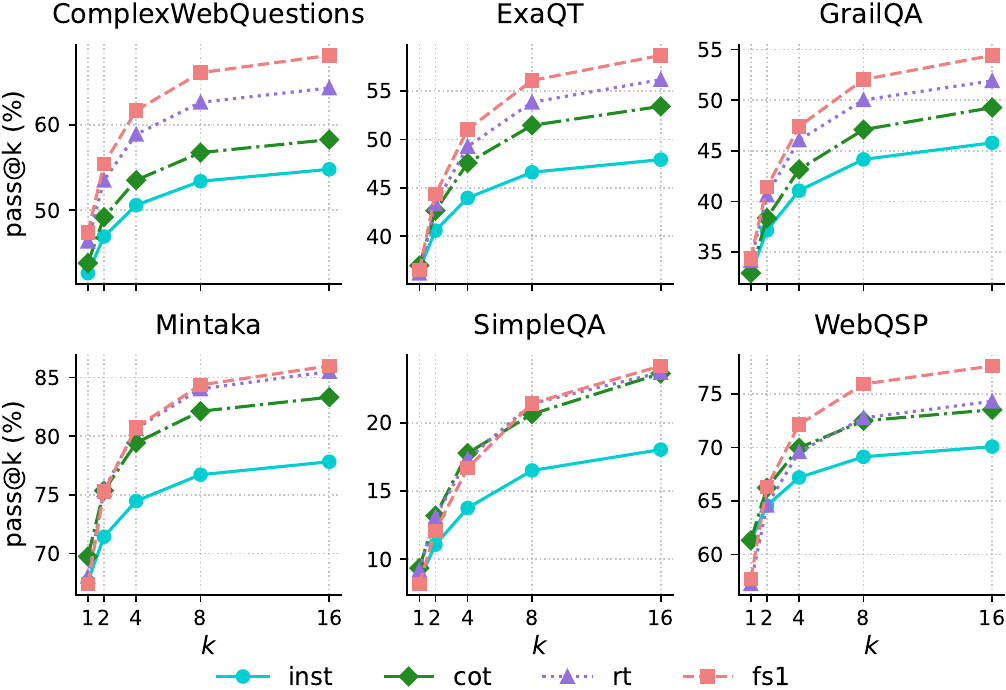}
    \caption{\textbf{Upper-bound Test-Time Scaling for Factual Reasoning.} We show with \texttt{Qwen2.5-32B} that parallel scaling is beneficial for complex mQA, measured by pass@$k$, especially when fine-tuned on \dataset{}, instead of conducting single-pass inference.}
    \label{fig:parallel-scaling-results}
\end{figure}

\section{Results}\label{results}


Parallel scaling can achieve lower latency by enabling multiple (identical) models to run simultaneously locally (via e.g., multiple GPUs or batching techniques) or via API based methods to generate multiple answers. 
Formally, parallel sampling entails an aggregation technique that combines $N$ independent solutions into a single final prediction, commonly known as a best-of-$N$ approach~\citep{chollet2019measure, irvine2023rewarding, brown2024large, li2022competition}. Formally, given a set of $N$ predictions $P = \{p_1, \dots, p_N\}$, the best-of-$N$ method selects a prediction $p \in P$ as the final output.

In this work, we present results using pass@$k$ (see~\cref{subsec:eval}), extending the number of sampled $k$ (until $k$ = 16). 
In \cref{fig:parallel-scaling-results}, we show parallel scaling results by performing 16 inference runs with \texttt{Qwen2.5-32B}-Instruct, CoT, \textbf{rt}, \dataset{} on each test dataset.\footnote{For parallel sampling, we limit ourselves to \texttt{Qwen2.5-32B} as running 16 inferences for 8 models for all 4 settings would require 12.2M model inferences for the test benchmarks, which is computationally prohibitive.}
As $k$ increases, pass@$k$ (indicating whether at least one generation is correct) rises steadily across all benchmarks. Parallel sampling boosts the chance of producing a correct answer, especially when fine-tuned on \dataset{}. 
For example, on CWQ, we see a performance increase of 16 absolute points at $k$ = 16 and on SimpleQA around 6 absolute points at the same $k$ compared to their original instruction-tuned counterpart.

\begin{figure}
    \centering
    \includegraphics[width=\linewidth]{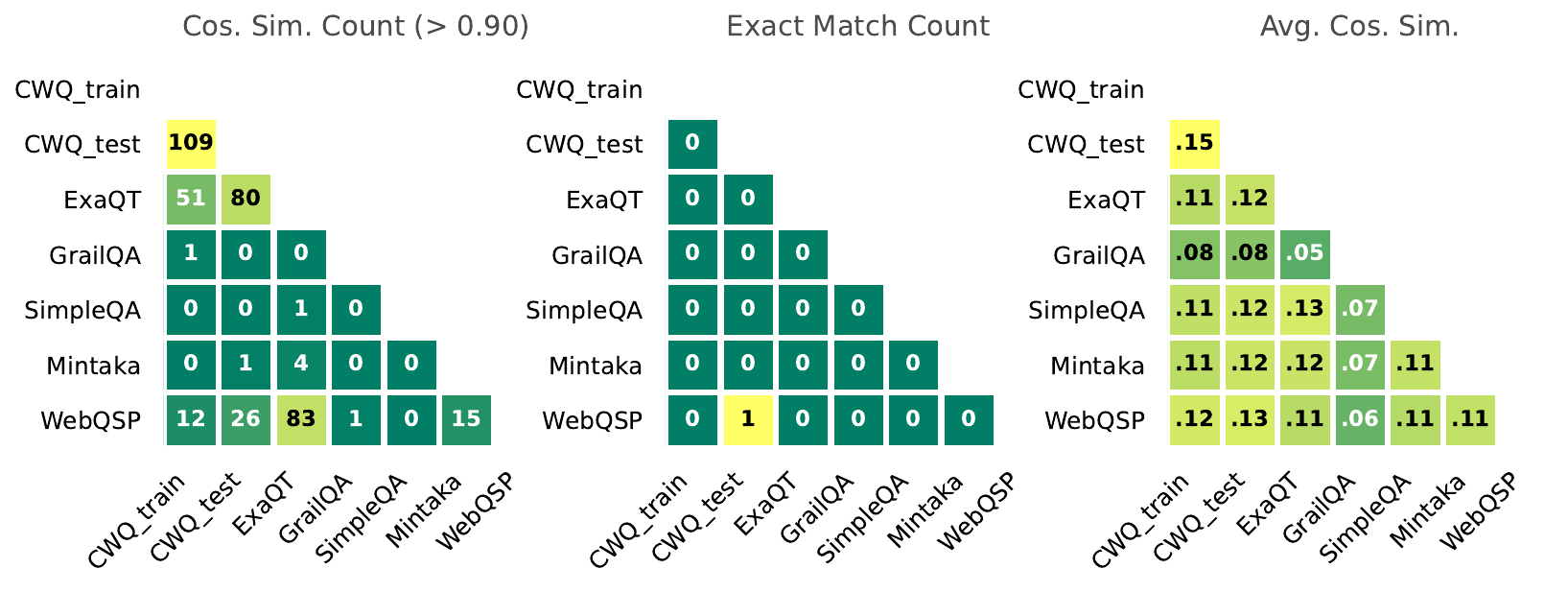}
    \caption{\textbf{Data Overlap.} We show data overlap between the train set and benchmark. On the left, one can observe the count of similar questions when the cosine similarity $>$0.90. In the middle, exact match counts. On the right, average pairwise cosine similarity.}
    \label{fig:overlap-analysis}
\end{figure}

\begin{table}[t]
    \centering
    \small
    \begin{tabular}{l|rr}
    \toprule
    \textbf{CWQ\_test}  &  \qwq{} & \texttt{R1-685B} \\
    \midrule
    pass@1     & 45.49 & 45.45 \\
    pass@2     & 54.90 & 54.91 \\
    pass@4     & 62.23 & 61.95 \\
    pass@8     & 67.79 & 67.41 \\
    pass@16    & 71.95 & 71.95 \\
    \bottomrule
    \end{tabular}
    \caption{\textbf{Ablation Teacher Model.} We show pass@$k$ performance of Qwen2.5-32B trained on separate subsets of \dataset{} separated by teacher model applied to CWQ. We show that there is almost no difference in performance. Indicating that \textbf{fs1} is the source of improvement.}
    \label{tab:ablation}
\end{table}

\begin{figure*}[t]
    \centering 
    \begin{minipage}[c]{0.64\textwidth}
    \begin{subfigure}[b]{0.43\linewidth}
        \centering
        \includegraphics[width=\linewidth]{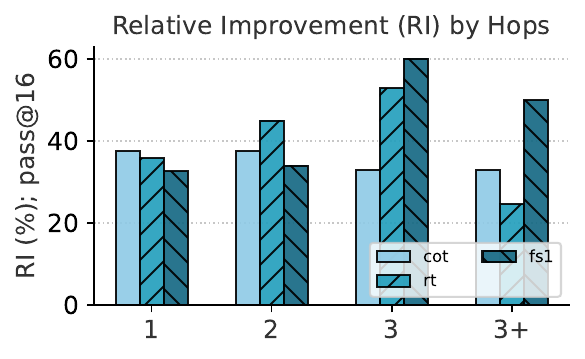}
        \vspace{-1.7em}
        \caption{Performance by number of hops required to answer the question.}
        \label{fig:hops}
    \end{subfigure}
    \begin{subfigure}[b]{0.53\linewidth}
        \centering
        \includegraphics[width=\linewidth]{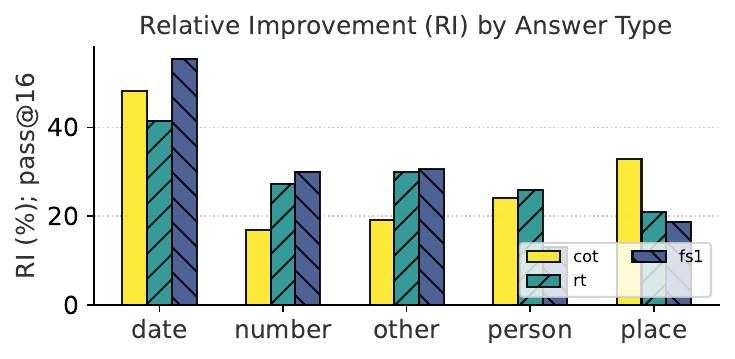}
        \vspace{-1.7em}
        \caption{Performance by answer type (i.e., what type of entity the answer is).}
        \label{fig:answer_type}
    \end{subfigure}\vspace{1em}
    \begin{subfigure}[b]{\linewidth}
        \centering
        \includegraphics[width=\linewidth]{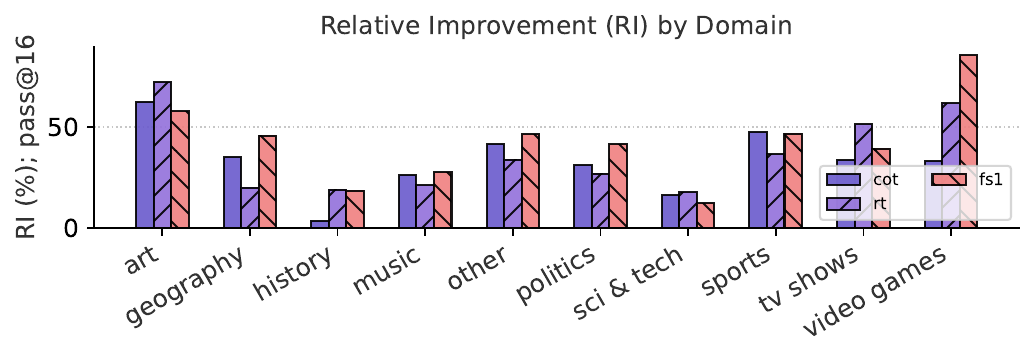}
        \vspace{-2em}
        \caption{Performance measured per domain.}
        \label{fig:domain}
    \end{subfigure}
    \end{minipage}\hfill
    \begin{minipage}[c]{0.26\textwidth}
    \caption{\textbf{Relative Improvements across Different Axes.} We show the relative performance improvement (\%) at pass@$16$ of different \texttt{Qwen-32B} (i.e., CoT, \textbf{rt} and \dataset{}) against the original instruct model. In (a), we show the performance of the models by the number of hops required to answer the question. In (b), we show the performance of the models by answer type. In (c), we show the performance by domain of the question. Absolute numbers are in~\cref{app:analysis_absolute} (\cref{app:analysis_absolute_section}).}
    \label{fig:analysis} 
    \end{minipage}
\end{figure*}

\section{Discussion and Ablations}

\subsection{Is the performance increase not just data leakage from the training set of fs1?}

In~\cref{fig:overlap-analysis}, we show the overlap of the questions in the train set of \texttt{ComplexWebQuestions} (CWQ\_train) versus all the other benchmarks used in our study (all Qs lower-cased). On the left, we count the times that the cosine similarity between questions exceeds 0.90. We can see that there is the most overlap between CWQ\_train and CWQ\_test (109 questions), and the second most is between WebQSP and ExaQT (83 questions). In the middle, we show that there is almost no exact string match between the questions. On the right, we show the average pairwise cosine similarity across the benchmarks is lower or equal to 0.15.

\begin{center}
\begin{tikzpicture}
\node [rq-box] (box){%
    \begin{minipage}{\textwidth}
         We find there is almost no data overlap between the main training set of \dataset{} and the benchmark test sets, indicating that the superior performance does not stem from data leakage and instead from \dataset{}.
    \end{minipage}
};
\node[rqbox-title,right=1em,below=-1em] at (box.north west) {\textbf{Finding §5.1}};
\end{tikzpicture}%
\end{center}

\subsection{Are the gains coming from a superior teacher model or \textbf{fs1}?}

It is not uncommon to assume that \qwq{} is a weaker model than \rone{} (685B). Therefore, it might be unclear how much the final performance gain comes from the superior reasoning of the teacher model or the factual grounding provided by the KG paths. To disentangle the effect of teacher model capabilities and \dataset{}, we take the subset of \qwq{} and \rone{} reasoning traces with the same questions (from~\cref{tab:data-statistics}) and fine-tune \texttt{Qwen2.5-32B} on the two \dataset{} subsets. In~\cref{tab:ablation}, we show that there is almost no difference in performance when using a different teacher model, when evaluated on CWQ, showing that \dataset{} works as a method by using KG paths to steer the reasoning behaviour of LLMs.

\begin{center}
\begin{tikzpicture}
\node [rq-box] (box){%
    \begin{minipage}{\textwidth}
         There is no difference between performance between different teacher models (i.e., \qwq{} vs.\ \texttt{R1-685B}), indicating that KG path injection into thinking traces helps downstream performance on mQA.
    \end{minipage}
};
\node[rqbox-title,right=1em,below=-1em] at (box.north west) {\textbf{Finding §5.2}};
\end{tikzpicture}%
\end{center}

\begin{table*}[t]
\centering
\definecolor{darkgreen}{HTML}{196F3D}
\definecolor{darkred}{HTML}{922B21}
\scriptsize
\begin{tabular}{lllllll}
\toprule
\textsc{Model} & \textbf{CWQ} & \textbf{ExaQT} & \textbf{GrailQA} & \textbf{SimpleQA} & \textbf{Mintaka} & \textbf{WebQSP} \\
\midrule
\multicolumn{7}{c}{\emph{Large Language Model Baselines}} \\
\midrule
Qwen2.5-72B & 0.481 & 0.440 & 0.361 & 0.117 & 0.736 & 0.653 \\
QwQ-32B & 0.479 & 0.390 & 0.358 & 0.097 & 0.708 & 0.612 \\
R1-70B & 0.501 & 0.476 & 0.340 & \textbf{0.188} & 0.755 & 0.549 \\
o3-mini & \textbf{0.558} & \textbf{0.497} & \textbf{0.438} & 0.138 & \textbf{0.774} & \textbf{0.680} \\
\midrule
\multicolumn{7}{c}{\emph{Small Language Models (0.36B-1.7B)}} \\
\midrule
SmolLM2-360M & 0.148 & 0.088 & 0.164 & 0.024 & 0.175 & 0.235 \\
\hspace{1em} + {cot} & 0.151 (\textcolor{darkgreen}{+2.0}\%) & 0.101 (\textcolor{darkgreen}{+14.8}\%) & \textbf{0.169} (\textcolor{darkgreen}{+3.0}\%) & 0.025 (\textcolor{darkgreen}{+4.2}\%) & 0.188 (\textcolor{darkgreen}{+7.4}\%) & 0.230 (\textcolor{darkred}{-2.1}\%) \\
\hspace{1em} + {\textbf{rt}} & \textbf{0.192} (\textcolor{darkgreen}{+29.7}\%) & \textbf{0.111} (\textcolor{darkgreen}{+26.1}\%) & 0.156 (\textcolor{darkred}{-4.9}\%) & \textbf{0.029} (\textcolor{darkgreen}{+20.8}\%) & \textbf{0.202} (\textcolor{darkgreen}{+15.4}\%) & \textbf{0.293} (\textcolor{darkgreen}{+24.7}\%) \\
\hspace{1em} + {\textbf{fs1}} & 0.179 (\textcolor{darkgreen}{+20.9}\%) & 0.093 (\textcolor{darkgreen}{+5.7}\%) & 0.138 (\textcolor{darkred}{-15.9}\%) & 0.027 (\textcolor{darkgreen}{+12.5}\%) & 0.197 (\textcolor{darkgreen}{+12.6}\%) & 0.264 (\textcolor{darkgreen}{+12.3}\%) \\
\midrule
Qwen2.5-0.5B & 0.135 & 0.058 & 0.127 & 0.023 & 0.131 & 0.173 \\
\hspace{1em} + {cot} & 0.161 (\textcolor{darkgreen}{+19.3}\%) & \textbf{0.104} (\textcolor{darkgreen}{+79.3}\%) & 0.141 (\textcolor{darkgreen}{+11.0}\%) & 0.031 (\textcolor{darkgreen}{+34.8}\%) & \textbf{0.214} (\textcolor{darkgreen}{+63.4}\%) & 0.234 (\textcolor{darkgreen}{+35.3}\%) \\
\hspace{1em} + {\textbf{rt}} & 0.190 (\textcolor{darkgreen}{+40.7}\%) & 0.089 (\textcolor{darkgreen}{+53.4}\%) & 0.155 (\textcolor{darkgreen}{+22.0}\%) & 0.022 (\textcolor{darkred}{-4.3}\%) & 0.178 (\textcolor{darkgreen}{+35.9}\%) & 0.286 (\textcolor{darkgreen}{+65.3}\%) \\
\hspace{1em} + {\textbf{fs1}} & \textbf{0.209} (\textcolor{darkgreen}{+54.8}\%) & 0.101 (\textcolor{darkgreen}{+74.1}\%) & \textbf{0.166} (\textcolor{darkgreen}{+30.7}\%) & \textbf{0.035} (\textcolor{darkgreen}{+52.2}\%) & 0.202 (\textcolor{darkgreen}{+54.2}\%) & \textbf{0.302} (\textcolor{darkgreen}{+74.6}\%) \\
\midrule
Qwen2.5-1.5B & 0.234 & 0.170 & 0.208 & 0.031 & 0.316 & 0.360 \\
\hspace{1em} + {cot} & 0.252 (\textcolor{darkgreen}{+7.7}\%) & \textbf{0.179} (\textcolor{darkgreen}{+5.3}\%) & \textbf{0.216} (\textcolor{darkgreen}{+3.8}\%) & \textbf{0.041} (\textcolor{darkgreen}{+32.3}\%) & \textbf{0.318} (\textcolor{darkgreen}{+0.6}\%) & \textbf{0.391} (\textcolor{darkgreen}{+8.6}\%) \\
\hspace{1em} + {\textbf{rt}} & 0.255 (\textcolor{darkgreen}{+9.0}\%) & 0.173 (\textcolor{darkgreen}{+1.8}\%) & 0.212 (\textcolor{darkgreen}{+1.9}\%) & 0.038 (\textcolor{darkgreen}{+22.6}\%) & 0.294 (\textcolor{darkred}{-7.0}\%) & 0.360 (\textcolor{darkgreen}{+0.0}\%) \\
\hspace{1em} + {\textbf{fs1}} & \textbf{0.263} (\textcolor{darkgreen}{+12.4}\%) & 0.162 (\textcolor{darkred}{-4.7}\%) & 0.204 (\textcolor{darkred}{-1.9}\%) & 0.035 (\textcolor{darkgreen}{+12.9}\%) & 0.301 (\textcolor{darkred}{-4.7}\%) & 0.356 (\textcolor{darkred}{-1.1}\%) \\
\midrule
SmolLM2-1.7B & 0.248 & 0.176 & 0.219 & 0.032 & 0.293 & 0.408 \\
\hspace{1em} + {cot} & 0.285 (\textcolor{darkgreen}{+14.9}\%) & 0.177 (\textcolor{darkgreen}{+0.6}\%) & 0.209 (\textcolor{darkred}{-4.6}\%) & 0.032 (\textcolor{darkgreen}{+0.0}\%) & 0.295 (\textcolor{darkgreen}{+0.7}\%) & 0.409 (\textcolor{darkgreen}{+0.2}\%) \\
\hspace{1em} + {\textbf{rt}} & \textbf{0.306} (\textcolor{darkgreen}{+23.4}\%) & \textbf{0.184} (\textcolor{darkgreen}{+4.5}\%) & \textbf{0.223} (\textcolor{darkgreen}{+1.8}\%) & \textbf{0.038} (\textcolor{darkgreen}{+18.7}\%) & \textbf{0.366} (\textcolor{darkgreen}{+24.9}\%) & \textbf{0.454} (\textcolor{darkgreen}{+11.3}\%) \\
\hspace{1em} + {\textbf{fs1}} & 0.305 (\textcolor{darkgreen}{+23.0}\%) & 0.179 (\textcolor{darkgreen}{+1.7}\%) & 0.218 (\textcolor{darkred}{-0.5}\%) & 0.036 (\textcolor{darkgreen}{+12.5}\%) & 0.341 (\textcolor{darkgreen}{+16.4}\%) & 0.426 (\textcolor{darkgreen}{+4.4}\%) \\
\midrule
\multicolumn{7}{c}{\emph{Large Language Models (3B-32B)}} \\
\midrule
Qwen2.5-3B & 0.317 & 0.214 & 0.252 & 0.044 & 0.396 & 0.466 \\
\hspace{1em} + {cot} & 0.302 (\textcolor{darkred}{-4.7}\%) & 0.222 (\textcolor{darkgreen}{+3.7}\%) & 0.248 (\textcolor{darkred}{-1.6}\%) & 0.048 (\textcolor{darkgreen}{+9.1}\%) & 0.431 (\textcolor{darkgreen}{+8.8}\%) & 0.477 (\textcolor{darkgreen}{+2.4}\%) \\
\hspace{1em} + {\textbf{rt}} & \textbf{0.363} (\textcolor{darkgreen}{+14.5}\%) & \textbf{0.235} (\textcolor{darkgreen}{+9.8}\%) & \textbf{0.279} (\textcolor{darkgreen}{+10.7}\%) & \textbf{0.053} (\textcolor{darkgreen}{+20.5}\%) & \textbf{0.495} (\textcolor{darkgreen}{+25.0}\%) & \textbf{0.483} (\textcolor{darkgreen}{+3.6}\%) \\
\hspace{1em} + {\textbf{fs1}} & 0.330 (\textcolor{darkgreen}{+4.1}\%) & 0.205 (\textcolor{darkred}{-4.2}\%) & 0.253 (\textcolor{darkgreen}{+0.4}\%) & 0.045 (\textcolor{darkgreen}{+2.3}\%) & 0.444 (\textcolor{darkgreen}{+12.1}\%) & 0.406 (\textcolor{darkred}{-12.9}\%) \\
\midrule
Qwen2.5-7B & 0.376 & 0.281 & 0.299 & \textbf{0.070} & 0.548 & \textbf{0.580} \\
\hspace{1em} + {cot} & 0.383 (\textcolor{darkgreen}{+1.9}\%) & 0.292 (\textcolor{darkgreen}{+3.9}\%) & 0.295 (\textcolor{darkred}{-1.3}\%) & 0.062 (\textcolor{darkred}{-11.4}\%) & \textbf{0.580} (\textcolor{darkgreen}{+5.8}\%) & 0.565 (\textcolor{darkred}{-2.6}\%) \\
\hspace{1em} + {\textbf{rt}} & 0.401 (\textcolor{darkgreen}{+6.6}\%) & \textbf{0.296} (\textcolor{darkgreen}{+5.3}\%) & 0.300 (\textcolor{darkgreen}{+0.3}\%) & 0.067 (\textcolor{darkred}{-4.3}\%) & 0.576 (\textcolor{darkgreen}{+5.1}\%) & 0.517 (\textcolor{darkred}{-10.9}\%) \\
\hspace{1em} + {\textbf{fs1}} & \textbf{0.408} (\textcolor{darkgreen}{+8.5}\%) & 0.272 (\textcolor{darkred}{-3.2}\%) & \textbf{0.303} (\textcolor{darkgreen}{+1.3}\%) & 0.053 (\textcolor{darkred}{-24.3}\%) & 0.551 (\textcolor{darkgreen}{+0.5}\%) & 0.492 (\textcolor{darkred}{-15.2}\%) \\
\midrule
Qwen2.5-14B & 0.392 & 0.336 & 0.318 & 0.068 & 0.624 & \textbf{0.599} \\
\hspace{1em} + {cot} & 0.422 (\textcolor{darkgreen}{+7.7}\%) & \textbf{0.356} (\textcolor{darkgreen}{+6.0}\%) & 0.322 (\textcolor{darkgreen}{+1.3}\%) & 0.080 (\textcolor{darkgreen}{+17.6}\%) & 0.664 (\textcolor{darkgreen}{+6.4}\%) & 0.592 (\textcolor{darkred}{-1.2}\%) \\
\hspace{1em} + {\textbf{rt}} & 0.451 (\textcolor{darkgreen}{+15.1}\%) & 0.352 (\textcolor{darkgreen}{+4.8}\%) & \textbf{0.331} (\textcolor{darkgreen}{+4.1}\%) & \textbf{0.082} (\textcolor{darkgreen}{+20.6}\%) & \textbf{0.678} (\textcolor{darkgreen}{+8.7}\%) & 0.562 (\textcolor{darkred}{-6.2}\%) \\
\hspace{1em} + {\textbf{fs1}} & \textbf{0.454} (\textcolor{darkgreen}{+15.8}\%) & 0.339 (\textcolor{darkgreen}{+0.9}\%) & 0.328 (\textcolor{darkgreen}{+3.1}\%) & 0.079 (\textcolor{darkgreen}{+16.2}\%) & 0.654 (\textcolor{darkgreen}{+4.8}\%) & 0.558 (\textcolor{darkred}{-6.8}\%) \\
\midrule
Qwen2.5-32B & 0.428 & 0.362 & 0.334 & 0.087 & 0.674 & \textbf{0.621} \\
\hspace{1em} + {cot} & 0.435 (\textcolor{darkgreen}{+1.6}\%) & \textbf{0.366} (\textcolor{darkgreen}{+1.1}\%) & 0.332 (\textcolor{darkred}{-0.6}\%) & \textbf{0.099} (\textcolor{darkgreen}{+13.8}\%) & \textbf{0.696} (\textcolor{darkgreen}{+3.3}\%) & 0.614 (\textcolor{darkred}{-1.1}\%) \\
\hspace{1em} + {\textbf{rt}} & 0.471 (\textcolor{darkgreen}{+10.0}\%) & \textbf{0.366} (\textcolor{darkgreen}{+1.1}\%) & 0.342 (\textcolor{darkgreen}{+2.4}\%) & 0.094 (\textcolor{darkgreen}{+8.0}\%) & 0.680 (\textcolor{darkgreen}{+0.9}\%) & 0.563 (\textcolor{darkred}{-9.3}\%) \\
\hspace{1em} + {\textbf{fs1}} & \textbf{0.477} (\textcolor{darkgreen}{+11.4}\%) & 0.361 (\textcolor{darkred}{-0.3}\%) & \textbf{0.344} (\textcolor{darkgreen}{+3.0}\%) & 0.078 (\textcolor{darkred}{-10.3}\%) & 0.682 (\textcolor{darkgreen}{+1.2}\%) & 0.576 (\textcolor{darkred}{-7.2}\%) \\
\bottomrule
\end{tabular}
\caption{\textbf{Single Pass (pass@1) Results on mQA Benchmarks.} We show accuracy and relative performance gains on our benchmarks for several baselines, Qwen2.5, and SmolLM2 models. For each size, we show the original instruction-tuned model followed by versions fine-tuned with CoT, \textbf{rt}, and \dataset{}. Parentheses indicate the relative improvement over the instruction-tuned counterpart. The benefits of fine-tuning are mostly for smaller models.}
\label{tab:passatone}
\end{table*}

\subsection{What type of samples do models tuned on \dataset{} seem to perform well on?}

In~\cref{fig:analysis}, we investigate what kind of questions the model (\texttt{Qwen2.5-32B}) seems to perform well on. 
We take metadata information from SimpleQA~\citep{wei2024measuring}, which indicates the question difficulty in number of hops required to answer the question (\cref{fig:hops}), type of answer (\cref{fig:answer_type}) and the domain of the question (\cref{fig:domain}). 
For question difficulty, we source the number of hops for each question in SimpleQA from~\citet{lavrinovics2025multihalmultilingualdatasetknowledgegraph}. 
We count the number of relations (P) from Wikidata, which would indicate the number of hops required to go from entity A to B. When a question does not contain any relations, we assume it takes more than 3 hops to answer the question.

In~\cref{fig:hops}, we observe that \dataset{} has lower relative improvements on easier questions (e.g., 1 or 2 hops required), but outperforms the other models when the question gets more complex (3 or more hops required). 
\emph{This indicates that inducing KG paths helps answering complex questions}.
In~\cref{fig:answer_type}, we show that \dataset{} has the most relative improvement on numerical answers, such as numbers, dates and also miscellaneous answer types.
Last, in~\cref{fig:domain}, \dataset{} performs best on questions related to video games, geography, politics, music, and miscellaneous questions. Additionally, \datasetrt{} performs best on art and history-related questions. Last, CoT performs best on questions related to sports.

\begin{center}
\begin{tikzpicture}
\node [rq-box] (box){%
    \begin{minipage}{\textwidth}
         We find that \dataset{} improves performance on more complex multi-hop questions (3+ hops), which indicates that including KG paths in thinking traces for training helps answering complex questions.
    \end{minipage}
};
\node[rqbox-title,right=1em,below=-1em] at (box.north west) {\textbf{Finding §5.3}};
\end{tikzpicture}%
\end{center}

\subsection{Single pass results across scale}
In~\cref{tab:passatone}, we show results in terms of accuracy via LLM-as-a-Judge at pass@$1$ (i.e., one inference run per question) on all test datasets.
For the baselines, we observe that {o3-mini} is the dominant model, achieving the highest score on five out of six datasets, such as its 0.774 accuracy on Mintaka and 0.680 on WebQSP. The only exception is SimpleQA, where {R1-70B} performs best with a score of {0.188}. These are followed by {Qwen2.5-72B-Instruct} and {QwQ-32B} in overall performance.

Observing the {Qwen2.5} results, the benefits of fine-tuning on \datasetrt{} and \dataset{} are most pronounced at the sub-billion parameter scale. For instance, fine-tuning the {0.5B} model on \dataset{} yields substantial relative gains across all tasks, peaking at a +74.6\% on WebQSP. However, as model size increases, the performance differences become more nuanced. For the {1.5B} model, the same \textbf{fs1} fine-tuning leads to performance degradation on four out of six datasets, such as ExaQT (-4.7\%) and WebQSP (-1.1\%). While larger models like the {32B} still benefit from fine-tuning (e.g., \textbf{rt} and \textbf{fs1} are often the best performers in their group), the relative gains are smaller than those seen at the 0.5B scale.

Our results also show that fine-tuning improvements do not uniformly generalize across different model families at the sub-billion parameter scale.
A comparison between the fine-tuned Qwen2.5 and SmolLM2 models reveals a significant performance divergence. Specifically, fine-tuning on \dataset{} provided consistent enhancements for the {Qwen2.5-0.5B} model, improving its CWQ score from 0.135 to 0.209. In contrast, the same fine-tuning on {SmolLM2-360M} yielded mixed results; while it improved performance on most tasks, it caused a notable degradation of -15.9\% on GrailQA.

This variance diminishes with scale, as models at the 1.5B/1.7B parameter scale exhibit more convergent behavior. For example, fine-tuning with \textbf{rt} on GrailQA provides a nearly identical small boost to both {Qwen2.5-1.5B} (+1.9\%) and \texttt{SmolLM2-1.7B} (+1.8\%). Overall, We hypothesize this scale-dependent effect may occur because larger models (e.g., 32B) possess stronger parametric knowledge, making them less reliant on the explicit guidance from KG paths.
\vspace{-0.5em}
\begin{center}
\begin{tikzpicture}
\node [rq-box] (box){%
    \begin{minipage}{\textwidth}
         We find that supervised fine-tuning with \datasetrt{} and \dataset{} especially helps smaller language models in terms of single pass accuracy and less for larger models compared to baselines such as CoT.
    \end{minipage}
};
\node[rqbox-title,right=1em,below=-1em] at (box.north west) {\textbf{Finding §5.4}};
\end{tikzpicture}%
\end{center}

\section{Related Work}\label{related-work}
Methods that involve long CoT processes~\citep{kojima2022large,wei2022chain} involving reflection, backtracking, \emph{thinking} (e.g.,~\citealp{deepseekai2025deepseekr1incentivizingreasoningcapability, muennighoff2025s1simpletesttimescaling}), self-consistency (e.g.,~\citealp{wang2022self}), and additional computation at inference time, such as test-time scaling~\citep{wu2024inference,muennighoff2025s1simpletesttimescaling,zhang2025whathowwherewell}, have shown promising improvements in LLM performance on complex reasoning tasks. Our work intersects with efforts in factuality, KG grounding, and test-time scaling.

\paragraph{Graph-enhanced In-context Learning and Reasoning.}
Enhancing the factual consistency of LLMs using KGs has been explored in different directions, including semantic parsing methods that convert natural language questions into formal KG queries~\citep{lanquery,ye2021rng}. QA-GNN~\cite{yasunaga-etal-2021-qa} enhances LMs by forming a joint graph of the question-answer context and a knowledge graph to perform mutual representation updates through graph-based message passing using graph neural nets (GNNs). Retrieval-augmented methods (KG-RAG)~\citep{li2023graph,jiang2023reasoninglm,sanmartin2024kg,sun2024thinkongraph,he2024g} aim to reduce LLMs' reliance on latent knowledge by incorporating structured information from a KG; reasoning on graphs (RoG) models~\citep{luo2023reasoning} generate relation paths grounded by KGs as faithful paths for the model to follow. \citet{sun2024thinkongraph} uses agents that iteratively performs a beam search on a knowledge graph, exploring, pruning, and reasoning over multiple paths until it determines enough information has been gathered to answer the question. G-Retriever~\citep{he2024g} is a RAG framework that answers questions about textual graphs by first retrieving a relevant, connected subgraph using a tree optimization formulation, and then generating a textual answer based on that subgraph. \citet{mavromatis-karypis-2025-gnn} might be the closest to our work, they use a GNN to process a dense subgraph and identify relevant answer candidate nodes, then retrieves the shortest paths connecting these candidates to the question entities, and finally provides these paths as verbalized context to an LLM for reasoning.

The earlier mentioned methods primarily focus on inference-time retrieval mechanisms, like iterative beam search~\citep{sun2024thinkongraph} or subgraph optimization~\citep{he2024g, mavromatis-karypis-2025-gnn}. Other works like~\citet{tan2025paths} also use KG paths to guide reasoning. Our work addresses another aspect: We focus on improving the model's intrinsic reasoning skill. Instead of inference-time retrieval, our method uses KGs as a one-time, offline process to create higher-quality training data, which induces the model to `think' more effectively.

\paragraph{Long Form Factuality.}
Factuality in NLP involves multiple challenges~\citep{augenstein2024factuality}, and while prior efforts have established reasoning datasets like SAFE~\citep{wei2024longform} and SimpleQA~\citep{wei2024measuring}, they often lack explicit grounding in structured knowledge subgraphs. In contrast, \citet{tian2024finetuning} directly address factual accuracy by fine-tuning models on automatically generated preference rankings that prioritize factual consistency. We train models directly on grounded factual data from knowledge graph to enhance factuality during LLM reasoning.

\paragraph{Test-Time Scaling as a Performance Upper-Bound.}
Our evaluation using pass@$k$ is situated within the broader context of test-time scaling, which seeks to improve performance by dedicating more compute at inference. This field encompasses parallel scaling (e.g., Best-of-N), where multiple candidate solutions are generated to increase the probability of finding a correct one~\citep{chollet2019measure, irvine2023rewarding, brown2024large, li2022competition}, and sequential scaling, where a single solution is iteratively refined through techniques like chain-of-thought prompting and revision~\citep{wei2022chain, nye2021show, madaan2024self, lee2025evolving, hou2025advancing, huang2022large, min2024imitate, muennighoff2025s1simpletesttimescaling, wang2024theoretical, li2025llms, jurayj2025finalanswertesttimescaling}. While practical applications of parallel scaling depend on a selection mechanism (e.g., majority voting or reward-model-based scoring) to choose the final answer~\citep{wang2022self, christiano2017deep, lightman2023letsverifystepstep, wang2024math, wu2024inference, beeching2024dvts, pan2024swegym, hassid2024larger, stroebl2024inference}, the performance of any such method is fundamentally limited by the quality of the underlying generations, often facing diminishing returns~\citep{brown2024largelanguagemonkeysscaling, snell2024scalingllmtesttimecompute, wu2024inference, levi2024simplemodelinferencescaling}. Our work focuses on improving the quality of each individual reasoning trace through fine-tuning, thereby directly boosting the upper-bound potential that is measured by pass@$k$.

\paragraph{Domain-specific Test-Time Scaling.}
Test-time scaling also spans specific domains like coding and medicine. Z1-7B optimizes coding tasks through constrained reasoning windows, reducing overthinking while maintaining accuracy~\citep{yu2025z1efficienttesttimescaling}. In medicine, extended reasoning boosts smaller models’ clinical QA performance significantly~\citep{huang2025m1unleashpotentialtesttime}, complemented by structured datasets like MedReason, which enhance factual reasoning via knowledge-graph-guided paths~\citep{wu2025medreasonelicitingfactualmedical}, similar to our work.

\section{Conclusion}\label{conclusion}
In this work, we have investigated whether grounding reasoning traces on knowledge graph paths and training models on them yield tangible gains in factual accuracy on complex open-domain QA tasks. 
After distilling over 3K original and knowledge-graph-enhanced reasoning traces from models \qwq{} and \rone{}, we fine-tuned 8 LLMs on \datasetrt{} and \dataset{} and evaluated them across 6 diverse benchmarks. 
In short, with parallel sampling, we consistently improve 6-14 absolute points in accuracy over their instruction-tuned counterpart. 
Particularly, using SimpleQA, we highlight that CoT and \datasetrt{} perform better on simpler questions (1 or 2 hops required), whereas our \dataset{}-tuned model performs better on more complex questions, requiring 3 hops or more. 
Lastly, we examined the performance of eight \dataset{}-tuned models across different parameter scales, finding that smaller models (below the 1.7B parameter range) show the largest increase in performance, while larger models see less profound improvements in a pass@1 setting. 
By releasing all code, models, and reasoning traces, we provide a rich resource for future work on process-level verification and the development of factuality-aware reward models. 
In turn, we hope this work facilitates more factual large language models, making them more useful for real-world usage.

\paragraph{Future Work.}
Future work could include scoring, e.g., \dataset{} reasoning traces to train process reward models that automatically evaluate intermediate reasoning steps, by grounding and scoring long-form generations with knowledge graph entries in each reasoning step to improve LLM factuality.

\section*{Ethics Statement}
The primary ethical motivation for this research is to enhance the factuality and reliability of LLMs. By addressing the probability of these models to generate incorrect information, our work aims to contribute positively to the development of more trustworthy AI systems. We do not foresee any direct negative ethical implications arising from this research. Instead, our goal is to provide a methodology that mitigates existing risks associated with misinformation, thereby promoting a safer and more beneficial application of language technologies.

\section*{Limitations}

Our approach assumes that conditioning on KG paths improves the accuracy of reasoning traces, though it does not guarantee perfect intermediate processes. Additionally, accurately evaluating entity answers poses challenges; we attempted to mitigate this limitation using LLM-based judgments, but these methods have their own inherent limitations. For evaluation, we note that pass@$k$ is an upper-bound performance measure. A practical implementation would require an additional selection mechanism, such as majority voting or a verifier model, to choose the final answer. Last, some of the test datasets used might be on the older side and English only, where we do not have control on whether the data has been included in any type of LLM pre- or post-training.

\section*{Acknowledgments}
We would like to thank the AAU-NLP group for helpful discussions and feedback on an earlier version of this article. 
MZ and JB were supported by the research grant (VIL57392) from VILLUM FONDEN.
MZ also received funding from the Danish Government to Danish Foundation Models (4378-00001B).
We acknowledge the Danish e-Infrastructure Cooperation for awarding this project access (No. 465001263; DeiC-AAU-N5-2024078 - H2-2024-18) to the LUMI supercomputer, owned by the EuroHPC Joint Undertaking, hosted by CSC (Finland) and the LUMI consortium through DeiC, Denmark.

\bibliography{custom,anthology-1,anthology-2}
\clearpage
\appendix

\section{Large Language Model Use}
We made use of LLMs to polish our writing and plotting our figures.

\section{Training and Inference}\label{app:trainandinference}
For running \rone{}, \texttt{o3-mini}, and some LLM-as-a-Judge experiments with \texttt{gpt-4o-mini}, we use API-based solutions via OpenAI\footnote{\url{https://platform.openai.com/}} or TogetherAI\footnote{\url{https://api.together.ai/}}. The costs of running inference on all data took around 250 USD. For fine-tuning and running inference of the local models, we make use of a large HPC cluster with hardware configurations comprising multiple nodes (depending on model size; e.g., 32B models require 4 nodes for training and 1 node for inference), each with node contains eight AMD MI250x GPU modules alongside a single 64-core AMD EPYC ``Trento'' CPU. The library we use for inference is \texttt{vllm}~\citep{kwon2023efficient} 0.9.3. For all the experiments it resulted in around 6,500 GPU hours spent. 

\subsection{Environmental Impact}
We acknowledge that conducting a large-scale analysis using LLMs comes with an environmental impact. Experiments were conducted using the LUMI cluster in Finland running on green energy. A cumulative of 6,500 GPU hours of computation was performed on AMD MI250x GPU modules, which has a TDP of 500 Watts. The experiments were ran from February to September 2025. During this time, the average carbon efficiency in Finland was 0.085 $kg/kWh$.\footnote{According to \url{https://app.electricitymaps.com/map}.} This means we released about 276 $kg$ of $CO_2$ equivalent. Estimations were conducted using the Machine Learning Impact calculator\footnote{Find the tool here: \url{https://mlco2.github.io/impact}.} presented in \citep{lacoste2019quantifying}.

\section{Example Reasoning Traces}\label{app:example}
We show four full examples in~\cref{fig:ex-reasoning-trace-r1} and \cref{fig:ex-reasoning-trace-qwq} for \textbf{rt} and \cref{fig:ex-reasoning-trace-fs1-qwq} and \cref{fig:ex-reasoning-trace-fs1-rone} for fs1. For visualization purposes, we split up the specific parts in special tags: \verb|<question>Q</question>|, \verb|<think>T</think>|, \verb|<gold>G</gold>| to indicate what it refers to.

\begin{figure*}[t]
\centering
\begin{tcolorbox}[title=Deepseek-R1 Example (\textbf{rt}), promptstyle]
\lstset{
    basicstyle=\normalfont\sffamily\scriptsize,
    breaklines=true,
    frame=none,
    columns=fullflexible,
}
\begin{lstlisting}[linewidth=\linewidth]
<question>
In Family Guy, Seth Macfarlane plays the parent of Chris Griffin. What is his character's name?
</question>

<think>
Okay, so I need to figure out the name of the character Seth MacFarlane voices who is the parent of Chris Griffin in Family Guy. Let me start by recalling what I know about the show.\n\nFirst, Family Guy is an animated sitcom created by Seth MacFarlane. He's known for voicing several characters in the show. The main family is the Griffins. The children in the family are Chris, Meg, and Stewie, and there's also the talking dog Brian. The parents would be the father and mother of these kids.\n\nI remember that the father is Peter Griffin, who is the bumbling, overweight patriarch of the family. Seth MacFarlane does voice Peter Griffin, along with other characters like Stewie and Brian. The mother is Lois Griffin, but I think she's voiced by Alex Borstein, not Seth MacFarlane. So, since the question is about the parent of Chris Griffin that Seth MacFarlane plays, it must be Peter Griffin.\n\nWait, let me double-check. Chris is the son, so his parents are Peter and Lois. Since Seth MacFarlane voices Peter, that's the answer. But just to be thorough, are there any other parents in the show that Seth voices? For example, maybe a grandparent? But Chris's grandparents would be Peter's parents or Lois's parents. I think Peter's father is named Francis Griffin, but I don't recall if Seth voices him. Lois's parents are Carter and Babs Pewterschmidt, and they are voiced by other actors. So no, the main parent characters voiced by Seth would be Peter. \n\nAlso, Stewie is the baby, and Brian is the dog, both voiced by Seth, but they aren't parents. So yes, the answer should be Peter Griffin.
</think>

The character voiced by Seth MacFarlane who is the parent of Chris Griffin in *Family Guy* is \\boxed{Peter Griffin}.

<gold>
["Peter Griffin", "Peter Lowenbrou Griffin"]
</gold>
\end{lstlisting}
\end{tcolorbox}
    \caption{\textbf{Example Instance (\textbf{rt}).} A successful reasoning trace for a simple multi-hop question by \rone{}.}
    \label{fig:ex-reasoning-trace-r1}
\end{figure*}

\begin{figure*}[t]
\centering
\begin{tcolorbox}[title=QwQ-32B Example (\textbf{rt}), promptstyle]
\lstset{
    basicstyle=\normalfont\sffamily\scriptsize,
    breaklines=true,
    frame=none,
    columns=fullflexible,
}
\begin{lstlisting}[linewidth=\linewidth]
<question>
What Eastern Europe main religion worshipers also attend St. Mary's Cathedral, in Batticaloa, Sri Lanka?
</question>

<think>
Okay, so the user is asking about the main religion worshippers in Eastern Europe who attend St. Mary's Cathedral in Batticaloa, Sri Lanka. Hmm, first I need to break this down. Let me start by recalling that Batticaloa is a city in Sri Lanka, which is in South Asia. The cathedral mentioned is a Christian church, specifically a Catholic cathedral since it's named St. Mary's.\n\nNow, the question is about Eastern Europeans who worship there. Wait, Eastern Europe's main religions are predominantly Christian, like Eastern Orthodox or Roman Catholic in some countries. But Sri Lanka's main religions are Buddhism, Hinduism, Islam, and then there's a Christian minority. So why would Eastern Europeans be attending a cathedral there? Maybe there's a community of Eastern Europeans living or working in Batticaloa? Or perhaps it's a question about the cathedral's affiliation?\n\nAlternatively, maybe there's confusion in the question. St. Mary's Cathedral in Batticaloa is likely part of the Roman Catholic Church. Eastern Europeans who are Catholic, like from countries such as Poland, Lithuania, or parts of Croatia, might attend. But Eastern Orthodox Christians would typically go to their own churches. Since the cathedral is Catholic, the main worshippers there would be local Catholics, but the question specifies Eastern Europeans. \n\nWait, perhaps the user is thinking of the term \"Eastern Europe\" in a different context. Maybe they're referring to the Eastern Catholic Churches, which are in communion with Rome but have Eastern rites. For example, Ukrainian Catholics or Ruthenian Catholics. But would they be in Sri Lanka? Maybe there's a small expat community or missionary group. Alternatively, maybe the question is mixing up terms, and the answer is Roman Catholic, as that's the cathedral's denomination, and Eastern Europeans of that faith would attend. \n\nAlternatively, could there be a confusion with Batticaloa's demographics? The area has a significant Tamil population, but Tamils are mostly Hindu or Christian. The cathedral serves the local Catholic community, which includes Sri Lankan Tamils and others. But the question specifically asks about Eastern Europeans. Maybe the answer is Roman Catholics from Eastern Europe, but I need to confirm if there's a specific group. \n\nAlternatively, perhaps the question is a trick, and the main religion in Eastern Europe (like Orthodox) isn't the one at the cathedral, so the answer is Roman Catholics. Since the cathedral is Catholic, Eastern European Catholics would be the ones attending. So the main religion worshippers there from Eastern Europe would be Roman Catholics. \n\nI think that's the most logical answer here. The cathedral is Catholic, so Eastern Europeans who are Roman Catholic would attend. The main religions in Eastern Europe include Orthodox and Catholic, so the Catholic ones from there would be the answer.
</think>

\\boxed{Roman Catholics}.

<gold>
["Catholicism", "Roman Catholic convert", "Roman Catholic"]
</gold>
\end{lstlisting}
\end{tcolorbox}
    \caption{\textbf{Example Instance (rt).} A successful reasoning trace for a simple multi-hop question by \qwq{}.}
    \label{fig:ex-reasoning-trace-qwq}
\end{figure*}

\begin{figure*}[t]
\centering
\begin{tcolorbox}[title=QwQ-32B Example (\textbf{fs1}), promptstyle]
\lstset{
    basicstyle=\normalfont\sffamily\scriptsize,
    breaklines=true,
    frame=none,
    columns=fullflexible,
}
\begin{lstlisting}[linewidth=\linewidth]
<question>
What geographic region is the country that used the Ottoman Lira used? 

While answering the question, make use of the following linearised graph as an inspiration in your reasoning, not as the only answer:

Konstantin Mikhailovich Bazili, place of birth, Istanbul
Istanbul, work location, Near East. 

Put your final answer within \boxed{}.
</question>

<think>
Okay, let's tackle this question step by step. The user is asking about the geographic region of the country that used the Ottoman Lira. First, I need to recall what the Ottoman Lira was. From what I remember, the lira was the currency used in the Ottoman Empire. So the country in question here is the Ottoman Empire. But since the Ottoman Empire is a historical entity and not a current country, the region would refer to where the empire was located. Now, the Ottoman Empire was a large and extensive empire that existed from the 14th century until after World War I. Its heartland was in Anatolia (modern-day Turkey) and Southeast Europe. The capital was Constantinople, which is now known as Istanbul, Turkey. The empire also controlled much of the Balkans, parts of North Africa, and the Middle East.

...
(cutoff)
...

Wait, but the question is about the geographic region of the country that used the Ottoman Lira. The country is the Ottoman Empire. The geographic region that it is in is the Near East. So combining the clues provided (Istanbul being part of the Near East), the answer would be the Near East. Alternatively, if they want the specific country now, like Turkey, but since the Ottoman Empire was not Turkey, but Turkey is the successor state. The question is a bit ambiguous. However, given the clues pointing towards Near East as the work location linked to Istanbul, I think the intended answer is the Near East region. So the final answer would be Near East.
</think>

The Ottoman Lira was the currency of the Ottoman Empire, which was centered in the geographic region known historically as the Near East. This region included the empire's heartland in Anatolia (modern-day Turkey), Istanbul (then Constantinople), parts of Southeast Europe, and the Middle East. The provided clues about Istanbul, the former capital, and its association with the Near East further reinforce this connection. 

\\boxed{Near East}.

<gold>
["Near East"]
</gold>
\end{lstlisting}
\end{tcolorbox}
    \caption{\textbf{Example Instance (fs1).} A successful reasoning trace for a question by \qwq{}.}
    \label{fig:ex-reasoning-trace-fs1-qwq}
\end{figure*}

\begin{figure*}[t]
\centering
\begin{tcolorbox}[title=Deepseek-R1 Example (\textbf{fs1}), promptstyle]
\lstset{
    basicstyle=\normalfont\sffamily\scriptsize,
    breaklines=true,
    frame=none,
    columns=fullflexible,
}
\begin{lstlisting}[linewidth=\linewidth]
<question>
What part of the world map is the Government of the Republic of Seychelles located in?
	
While answering the question, make use of the following linearised graph as an inspiration in your reasoning, not as the only answer:

Seychellois Creole, country, Seychelles
Seychelles, continent, Africa
Seychellois Creole, indigenous to, Seychelles
Seychelles, continent, Africa. 

Put your final answer within \boxed{}.
</question>

<think>
Okay, so I need to figure out where the Government of the Republic of Seychelles is located on the world map. Let me start by recalling what I know about Seychelles. I think Seychelles is a country made up of islands. But where exactly? The user provided a linearised graph that mentions Seychelles is in the continent of Africa. Hmm, that's interesting because when I think of Africa, I usually picture the mainland, but there are also island countries associated with Africa. Let me verify that. Seychelles is an archipelago in the Indian Ocean, right? So it's a group of islands northeast of Madagascar. Since Madagascar itself is off the southeastern coast of Africa, Seychelles being northeast of there would place it in the Indian Ocean, near the African continent. The graph also mentions Seychellois Creole being indigenous to Seychelles, which makes sense as the local language. So, putting this together, the Government of the Republic of Seychelles is located in the country of Seychelles, which is part of the African continent. Even though it's an island nation, geographically and politically, it's considered part of Africa. Therefore, the answer should be that the government is located in Seychelles, which is in Africa.
</think>

The Government of the Republic of Seychelles is located in the country of Seychelles. According to the provided graph, Seychelles is part of the continent of Africa. Although geographically situated in the Indian Ocean northeast of Madagascar, it is politically and regionally classified as part of Africa. 

\\boxed{Africa}

<gold>
[ "Africa", "Afrique" ]
</gold>
\end{lstlisting}
\end{tcolorbox}
    \caption{\textbf{Example Instance (fs1).} A successful reasoning trace for a question by \rone{}.}
    \label{fig:ex-reasoning-trace-fs1-rone}
\end{figure*}

\section{Test Benchmark}\label{app:test-benchmark}

\subsection{Examples Test Benchmark}\label{app:test-examples}
In~\cref{fig:test-examples}, we show for each dataset in the test benchmark an example of how the question and answers look like.

\begin{figure*}[t]
\centering
\begin{tcolorbox}[title=Benchmark Test Examples, promptstyle]
\lstset{
    basicstyle=\normalfont\sffamily\scriptsize,
    breaklines=true,
    frame=none,
    columns=fullflexible,
}
\begin{lstlisting}[linewidth=\linewidth]
<cwq>
Who was the president in 2012 of the country where Unidad de Valor Constante is the chosen currency?
Answers: [ "Rafael Correa", "Rafael Correa Delgado", "Rafael Vicente Correa Delgado" ]
</cwq>

<exaqt>
what nba team won the back to back championship in 1989 and 1990?
Answer: [ "Detroit Pistons" ]
</exaqt>

<grailqa>
lonnie wheeler contributed to a book edition published by what publisher?
Answer: [ "Simon & Schuster" ]
</grailqa>

<simpleqa>
Who received the IEEE Frank Rosenblatt Award in 2010?
Answer: [ "Michio Sugeno" ]
</simpleqa>

<mintaka>
How many books are in Goosebumps?
Answer: [ "235" ]
</mintaka>

<webqsp>
where did diego velazquez die?
Answer: [ "Madrid" ]
</webqsp>
\end{lstlisting}
\end{tcolorbox}
    \caption{\textbf{Text Examples.} For each dataset in the benchmark, we show an example.}
    \label{fig:test-examples}
\end{figure*}

\section{SPARQL queries}\label{app:sparql}
The query format to retrieve the Wikidata entities for Freebase entities is given by

\begin{framed}
\begin{Verbatim}[fontsize=\small]
SELECT ?wikientity 
WHERE
{?wikientity wdt:P646 $FREEBASE_ENTITY}
\end{Verbatim}
\end{framed}

\noindent The general structure of the SPARQL queries for 2-hop paths between the source and target entities are given by

\begin{framed}
\begin{Verbatim}[fontsize=\scriptsize]

SELECT ?p1 ?p1Label ?o1 ?o1Label ?p2 ?p2Label 
WHERE
{wd:$SOURCE_ENTITY ?p1  ?o1.
?o1 ?p2 wd:$TARGET_ENTITY}
SERVICE wikibase:label { bd:serviceParam wikibase:language 
"[AUTO_LANGUAGE],mul,en". }
\end{Verbatim}
\end{framed}

\begin{framed}
\begin{Verbatim}[fontsize=\scriptsize]

SELECT ?p1 ?p1Label ?o1 ?o1Label ?p2 ?p2Label 
WHERE 
{wd:$TARGET_ENTITY ?p1 ?o1. 
?o1 ?p2 wd:$SOURCE_ENTITY} 
SERVICE wikibase:label { bd:serviceParam wikibase:language 
"[AUTO_LANGUAGE],mul,en". }
\end{Verbatim}
\end{framed}

\section{Absolute Performance Results for Analysis}\label{app:analysis_absolute_section}

\begin{figure*}[t]
    \centering 
    \begin{subfigure}[b]{0.4\linewidth}
        \centering
        \includegraphics[width=\linewidth]{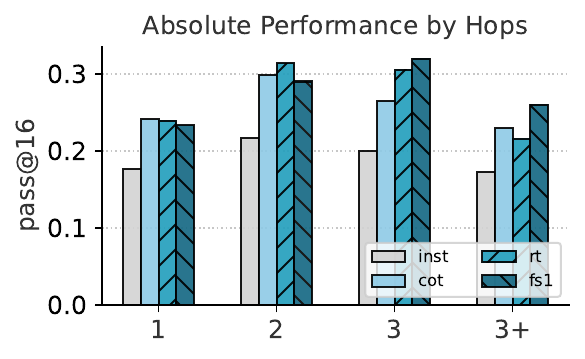}
        \caption{Performance by number of hops required to answer the question measured in pass@$16$.}
        \label{fig:hops_absolute}
    \end{subfigure}\hspace{1em}
    \begin{subfigure}[b]{0.5\linewidth}
        \centering
        \includegraphics[width=\linewidth]{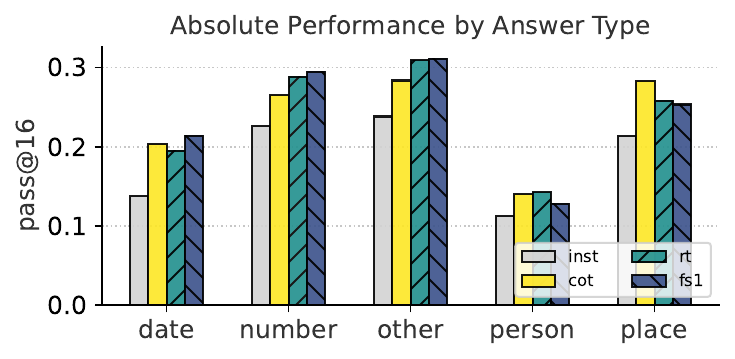}
        \caption{Performance by answer type (i.e., what type of entity the answer is) in pass@$16$.}
        \label{fig:answer_type_absolute}
    \end{subfigure}\hspace{1em}
    \begin{subfigure}[b]{.95\linewidth}
        \centering
        \includegraphics[width=\linewidth]{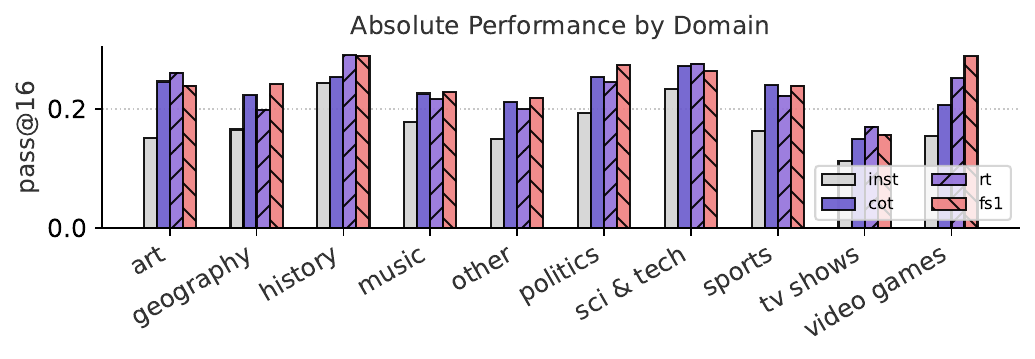}
        \caption{Performance measured per domain in pass@$16$.}
        \label{fig:domain_absolute}
    \end{subfigure}
    \caption{\textbf{Absolute Performance across Different Axes.} We show the absolute performance at pass@$16$ of different versions of \texttt{Qwen-32B} (i.e., instruct, CoT, \textbf{rt} and \dataset{}). In (a), we show the performance of the models by answer type. In (b), we show the performance of the models by the number of hops required to answer the question. In (c), we show the performance by domain of the question.}
    \label{app:analysis_absolute} 
\end{figure*}

We show the absolute performance numbers from~\cref{fig:analysis} at pass@$16$ in \cref{app:analysis_absolute}.

\end{document}